\pdfoutput=1

\documentclass[11pt]{article}
\usepackage{authblk}

\usepackage[]{emnlp2021}

\usepackage{times}
\usepackage{latexsym}

\usepackage[T1]{fontenc}

\usepackage[utf8]{inputenc}

\usepackage{microtype}

\usepackage{graphicx}
\usepackage{multirow}
\usepackage{arydshln}

\title{Effect of Visual Extensions on Natural Language Understanding in Vision-and-Language Models}

\author[1,2]{{\bf Taichi Iki}}
\author[1,2]{{\bf Akiko Aizawa}}
\affil[1]{National Institute of Informatics, Chiyoda-ku, Tokyo, Japan}
\affil[2]{Graduate University for Advanced Studies, Hayama, Kanagawa, Japan}
\affil[ ]{\texttt{\{iki,aizawa\}@nii.ac.jp}}

\begin{document}
\maketitle
\begin{abstract}
A method for creating a vision-and-language (V\&L) model is to extend a language model through structural modifications and V\&L pre-training. 
Such an extension aims to make a V\&L model inherit the capability of natural language understanding (NLU) from the original language model.
To see how well this is achieved, we propose to evaluate V\&L models using an NLU benchmark (GLUE). 
We compare five V\&L models, including single-stream and dual-stream models, trained with the same pre-training.
Dual-stream models, with their higher modality independence achieved by approximately doubling the number of parameters, are expected to preserve the NLU capability better.
Our main finding is that the dual-stream scores are not much different than the single-stream scores, contrary to expectation. 
Further analysis shows that pre-training causes the performance drop in NLU tasks with few exceptions.
These results suggest that adopting a single-stream structure and devising the pre-training could be an effective method for improving the maintenance of language knowledge in V\&L extensions.
\end{abstract}

\section{Introduction}

Pre-trained vision-and-language (V\&L) models improve the performance of tasks that require an understanding of the V\&L grounding, including visual question answering~\cite{antol2015vqa}, referring expression comprehension~\cite{KazemzadehOrdonezMattenBergEMNLP14}, and image-text matching (ITM)~\cite{suhr-etal-2019-corpus}.
Recent V\&L tasks, such as multi-modal reading comprehension~\cite{kembhavi2017you,yagcioglu-etal-2018-recipeqa,hannan2020manymodalqa,VisualMRC2021} and dialogue~\cite{ilinykh2019meetup,haber2019photobook,udagawa2019natural}, require a deeper NLU as well as the grounding. 
Extending pre-trained language models (LMs) is an option for those tasks as this allows V\&L models to inherit language knowledge from their source LMs.
The typical extending consists of visual pre-training and structure such as the stream type; the single-stream inserts vision tokens into the input sequence of the LM, and the dual-stream uses another sequence for early visual encoding.

One of the remaining challenges is to understand how such extensions affect the pre-trained language knowledge.
For example, \citet{NEURIPS2019_c74d97b0} proposed the dual-stream model where part of the goal was to protect the learned LMs.
The authors focused on evaluation with V\&L tasks and did not evaluate their models with language-only tasks.
\citet{cao2020behind} evaluated the extent of language knowledge loss in the single/dual-stream models against the source LM using language-only tasks.
However, the difference between single-stream and dual-stream models was unclear because the pre-training was also different in their models.

In this paper, we investigate the effect of visual extensions in V\&L models on language-only tasks\footnote{The source code for our experiments is available at \url{https://github.com/Alab-NII/eval_vl_glue}}.
\citet{bugliarello2020multimodal} proposed a framework to unify transformer-based V\&L models and compared some single/dual-stream models in the same setup. 
Based on their work, our study shows how these structural differences affect the performance of NLU using the GLUE~\cite{wang2018glue} tasks.

In our experiments, fine-tuning of pre-trained V\&L models shows that both single/dual-stream models perform worse than the source LM and that single-stream models perform slightly better than dual-stream models.
Further, we fine-tune the models created by only structural modifications without pre-training.
We observe that the single/dual modification alone has little effect on the GLUE scores, indicating the performance degradation is primarily caused by pre-training. 
We also see how the V\&L models changed from the source LM by analyzing the changes in the model parameters and the problem sets that each model can solve.
Our results suggest that it would be more effective to adopt a single stream, and devise pre-training strategies for maintaining language knowledge.

\begin{table*}
\small
\centering
\scalebox{0.94}{
\begin{tabular}{cccccccc}
\hline
\bf{Structure} & \begin{tabular}[c]{@{}c@{}} \bf{Abbreviation} \\ \bf{in this paper} \end{tabular} & \bf{Stream} & \bf{\#param} & \begin{tabular}[c]{@{}c@{}} \bf{Location} \\ \bf{format} \end{tabular} & \begin{tabular}[c]{@{}c@{}} \bf{Global} \\ \bf{image feat.} \end{tabular} & \begin{tabular}[c]{@{}c@{}} \bf{Vision} \\ \bf{type ID} \end{tabular} & \begin{tabular}[c]{@{}c@{}}\bf{Original} \\ \bf{paper} \end{tabular}  \\ \hline

VisualBERT$_{\rm CTRL}$ & 
VIS$_{\rm CTRL}$ & 
\multirow{4}{*}{Single} & 
112M & 
not used & 
head & 
from BERT &
\citet{li2019visualbert} \\

Uniter$_{\rm CTRL}$ & 
UNI$_{\rm CTRL}$ &
&
112M &
LTRBA &
head &
from BERT &
\citet{chen2020uniter} \\

VL-BERT$_{\rm CTRL}$ & 
VL$_{\rm CTRL}$ &  
& 
114M & 
LTRBA & 
\begin{tabular}[c]{@{}c@{}}head + added \\ to each word\end{tabular} & 
extended &
\citet{Su2020VL-BERT:} \\ \hdashline

ViLBERT$_{\rm CTRL}$ & 
VIL$_{\rm CTRL}$ & 
\multirow{2}{*}{Dual} & 
240M & 
LTRBA & 
head & 
not used & 
\citet{NEURIPS2019_c74d97b0} \\

LXMERT$_{\rm CTRL}$ & 
LX$_{\rm CTRL}$ & 
\multicolumn{1}{l}{} & 
209M & 
LTRBA & 
head & 
not used & 
\citet{tan-bansal-2019-lxmert} \\ \hline
\end{tabular}
}
\vspace{-2mm}
\caption{Comparison of the structures in the {\bf controlled} setup used in this study.
L, T, R, B, and A in the location format column denote left, top, right, bottom, and area, respectively.
\citet{bugliarello2020multimodal}'s controlled setup unifies the use of the location format and global visual features, which were different in the original proposals.
\label{table:structure}
}
\vspace{-3mm}
\end{table*}

\section{Controlled V\&L Models}

In this section, we describe the pre-trained V\&L models used in our experiments.
\citet{bugliarello2020multimodal} proposed a framework for V\&L models that consider a sequence of tokens in sentences as language information, and a sequence of recognized object regions as visual information.
In their framework, they reproduced five existing models, VisualBERT~\cite{li2019visualbert}, Uniter~\cite{chen2020uniter}, VL-BERT~\cite{Su2020VL-BERT:}, ViLBERT~\cite{NEURIPS2019_c74d97b0}, and LXMERT~\cite{tan-bansal-2019-lxmert}, and made their controlled versions by modifying some parts for a fairer and easier comparison.
We use these controlled versions.

\subsection{Structural Modification}
We describe streams and embeddings, which are the basic factors of the model structures.
We summarize the model structures in the controlled setup used in this experiment in Table~\ref{table:structure}.

\paragraph{Streams.} 
V\&L models can be divided into two categories based on how the vision and language sequences are encoded.
Single-stream models, VisualBERT, Uniter, and VL-BERT, jointly process the vision and language sequences in a single encoder.
Dual-stream models, ViLBERT and LXMERT, encode those sequences separately before encoding them jointly.
ViLBERT is an early example of the dual-stream models
and was proposed mainly to account for the differences in abstraction levels between vision and language, and to protect learned language models.
In the controlled setup of \citet{bugliarello2020multimodal}, the stream type is identical to the original one in all models.

\paragraph{Embeddings.}
The major difference in embeddings is the use of global visual feature.
The original VisualBERT, Uniter, and LXMERT do not use the global visual feature.
ViLBERT has a token that represents the global visual feature at the beginning of vision sequences.
VL-BERT inserts the global visual feature to the last of vision sequences and also adds the global visual feature to each token embedding in the language sequence.
Object location is also expressed differently.
The original VL-BERT and LXMERT use four attributes (left, top, right, bottom).
In addition to the four attributes, the original ViLBERT uses area, and the original Uniter uses width, height, and area.
VisualBERT does not use location information\footnote{If alignments between words and regions are provided, VisualBERT adds the same position embeddings to matched word and region tokens instead.}.

The controlled setup is based on the structure of ViLBERT.
For the global image feature, the setup inserts the average of vision tokens to the head of the vision sequence for all models. 
In addition to inserting the global visual feature, the controlled VL-BERT adds it to the respective tokens in the language sequence.
For location, VisualBERT's setup that do not use location information remain the same, while the other models use the five attributes. 
The five attributes are normalized by width or height.
Another point is the token type for the vision tokens.
In the controlled setup, the token type is not added for ViLBERT and LXMERT because they have separate streams.
Of the single-stream models, VisualBERT and Uniter use BERT's token type ID to specify vision tokens, while VL-BERT adds a new embedding to represent vision tokens.

\subsection{Pre-training}
We summarize the pre-training used in the controlled setup to train the five model structures described above.
Note that we omit the detail of the pre-training used in each original paper here.

The five models were pre-trained on Google’s Conceptual Captions~\cite{sharma-etal-2018-conceptual} corpus, which was collected from Web images and their alt-text HTML attributes.
The corpus was filtered before training, and the size was approximately 2.7 M pairs as a result.
Three tasks, masked language modelling~(MLM), masked object classification~(MOC), and ITM, were made from image-text pairs in the corpus. 
Given an image-text pair, the model predicts masked language tokens for MLM, the object class of masked vision tokens for MOC, and whether the pair is correct or not for ITM.

The weights of the five models were initialized with the pre-trained weights of BERT$_{\rm BASE}$ if the corresponding weights were in BERT$_{\rm BASE}$; otherwise (e.g., the weights of the vision encoder in dual-stream models), they were initialized randomly.

\section{Experiment with GLUE}

\subsection{Datasets}

The GLUE benchmark \cite{wang2018glue} is a collection of diverse tasks for studying NLU systems.
It consists of nine tasks: CoLA~\cite{warstadt-etal-2019-neural}, SST-2~\cite{socher2013recursive}, MRPC~\cite{dolan2005automatically}, QQP\footnote{\url{https://www.kaggle.com/c/quora-question-pairs}}, STS-B~\cite{cer2017semeval}, MNLI~\cite{nangia2017repeval}, QNLI~\cite{rajpurkar2016squad}, RTE~\cite{dagan2006pascal,bar2006second,giampiccolo2007third,bentivogli2009fifth}, and WNLI~\cite{levesque2011winograd}.
STS-B is a single-valued regression task, and the others are classification tasks.
We train the controlled pre-trained models on the training sets and
evaluate them with the development sets.
Figure~\ref{figure:task_scores} (left) shows the number of the training sentences in the corpora and their word overlap between the corpus used in the V\&L pre-training.

\subsection{Implementation Details}
We fine-tuned pre-trained models published by \citet{bugliarello2020multimodal}~\footnote{
\url{https://github.com/e-bug/volta/blob/main/MODELS.md}
}.
To use a script for the GLUE benchmark, we modified the model codes for Huggingface's Transformers~\cite{wolf-etal-2020-transformers}\footnote{We checked that our implementation reproduced original results with a V\&L task, NLVR$^2$~\cite{suhr-etal-2019-corpus}.}.
We used the BERT-uncased tokenizer to tokenize sentences.

\paragraph{Image inputs.} 
Because the GLUE tasks have no image input, we used a black image~(of $224 \times 224$ pixels) in our experiments.
We followed the method of \citet{bugliarello2020multimodal} for image processing;
we input the images to the Faster R-CNN object detector~\cite{ren2016faster} trained for the Bottom-Up and Top-Down model~\cite{anderson2018bottom}, and used the top 36 detected results (bounding boxes and feature vectors) as vision tokens\footnote{Although the image was monochromatic black, 36 bounding boxes with different features were detected.}.
We used the average of the vision tokens as a global visual token.
Those vision tokens were fixed and used for both training and evaluation in all models.

In this study, we tried image completion with black images for tasks where no image is provided as a simple way to preserve the input format used in pre-training.
However, there are many possible methods for complementing the image input.
For example, a method as simple as the present one can use other images, a noise input, or learnable parameters.
Examining the impact of image input completion methods remains as future work.

\paragraph{Head for classification.}
We adopted the method used in \citet{bugliarello2020multimodal} for V\&L tasks.
We used a learnable linear layer to calculate the likelihood of document classes, such as entailment/neutral/contradiction.
We input the element-wise product of two vectors made from the model's output sequence into the linear layer.
For those two vectors, we pooled the portions of the model's output sequence that correspond to the vision input and to the language input, respectively, by taking the first token of each portion.
This corresponds to taking the outputs of the [CLS] token (in the language sequence) and the global visual token.

\paragraph{Hyperparameters for fine-tuning.} 
We used a batch size of 64 and Adam for optimization. 
The learning rate was initialized at 2e-5 and decreased linearly. 
We trained for five epochs, evaluating the loss on the dev sets at the end of each epoch. 
Finally, we adopted the model with the lowest loss.

\subsection{Overall Result}

\begin{table*}[]
\small
\centering
\scalebox{0.93}{
\begin{tabular}{lcccccccccc|c}
\hline
 & \multicolumn{9}{c}{GLUE~(Language)} & ~ & V\&L \\
\textbf{} & \textbf{avg$\uparrow$} (SD) & \textbf{CoLA} & \textbf{SST-2} & \textbf{MRPC} & \textbf{QQP} & \textbf{STS-B} & \textbf{MNLI} & \textbf{QNLI} & \textbf{RTE} & \textbf{WNLI} & \textbf{avg$\uparrow$} \\ \hline
BiLSTM & 66.7 & 17.6 & 87.5 & 77.9/85.1 & 85.3/82.0 & 71.6/72.0 & 66.7 & 77.0 & 58.5 & 56.3 & ~ \\
BERT$_{\rm BASE}$ & 77.3 (0.8) & 54.6 & 92.5 & 81.9/87.6 & 90.6/87.4 & 88.2/87.9 & 84.4 & 91.0 & 62.5 & 48.8 & ~ \\ \hline
Model avg & 71.6 & 38.0 & 88.9 & 70.1/80.8 & 89.0/85.5 & 78.5/78.7 & 81.0 & 85.5 & 55.7 & 52.6 & 68.0 \\ \hdashline
\hspace{2mm} VIS$_{\rm CTRL}$ & \textbf{72.5} (1.2) & 38.6 & 89.4 & \textbf{71.9/82.1} & \textbf{89.4/86.0} & 81.8/81.7 & \textbf{81.8} & \textbf{87.0} & 56.6 & 53.1 & 69.2 \\
\hspace{2mm} UNI$_{\rm CTRL}$ & 71.4 (0.3) & 37.4 & 89.7 & 69.3/80.3 & 89.2/85.7 & 74.9/75.6 & 81.2 & 86.0 & 55.6 & \textbf{55.4} & \textbf{69.7} \\
\hspace{2mm} VL$_{\rm CTRL}$& \textbf{72.4} (0.8) & 38.7 & 89.8 & 70.6/81.8 & 89.0/85.4 & \textbf{82.9/82.8} & 81.4 & 86.3 & 55.7 & 53.1 & 67.7 \\ \hdashline
\hspace{2mm} VIL$_{\rm CTRL}$ & 70.9 (0.8) & 36.1 & \textbf{90.4} & 69.0/79.4 & 88.6/85.0 & 77.7/78.0 & 80.1 & 83.8 & 53.7 & \textbf{55.4} & \textbf{69.8}\\
\hspace{2mm} LX$_{\rm CTRL}$ & 70.5 (0.2) & \textbf{39.0} & 90.2 & 69.8/80.4 & 89.0/85.4 & 75.3/75.3 & 80.7 & 84.2 & \textbf{57.2} & 46.0 & 63.6\\ \hline
\end{tabular}
}
\vspace{-2mm}
\caption{Performance of the development sets of the GLUE tasks (single-task training). 
The best scores among the five V\&L models are shown in bold.
We report the Matthews correlation for CoLA; accuracy/F1 for MRPC and QQP; the Pearson/Spearman correlation for STS-B; and the accuracy for all other tasks.
For MNLI, we show accuracy averaged over the matched and mismatched sets.
The values of BiLSTM are cited from \citet{wang2018glue}.
The other values related to GLUE are our results.
We fine-tuned the pre-trained models for each task three times with different random seeds.
We show the standard deviation in parentheses for avg and in Appendix~\ref{sec:appendix_overall} for each task.
In the last column, we also show the scores of V\&L tasks calculated by averaging the results in \citet{bugliarello2020multimodal}.
The detail is described in Section~\ref{sec:l_and_vl}.
\label{table:overall}
}
\vspace{-3mm}
\end{table*}

Table~\ref{table:overall} shows the results of the GLUE benchmark.
In our experiment, we fine-tuned five V\&L models and their source language model--BERT$_{\rm BASE}$.
We also cited the BiLSTM baseline from the GLUE paper.
The Glue avg of five V\&L models decrease compared to BERT$_{\rm BASE}$.
We can see a trend where the single-stream models perform slightly better than the dual-stream models.
Note that this trend is consistent with the results of \citet{cao2020behind} for linguistic probing of the original Uniter and LXMERT.
Although the difference is small, this suggests that the single-stream models can maintain more of BERT$_{\rm BASE}$'s knowledge.

\begin{figure}[]
\centering
\includegraphics[width=0.5\linewidth]{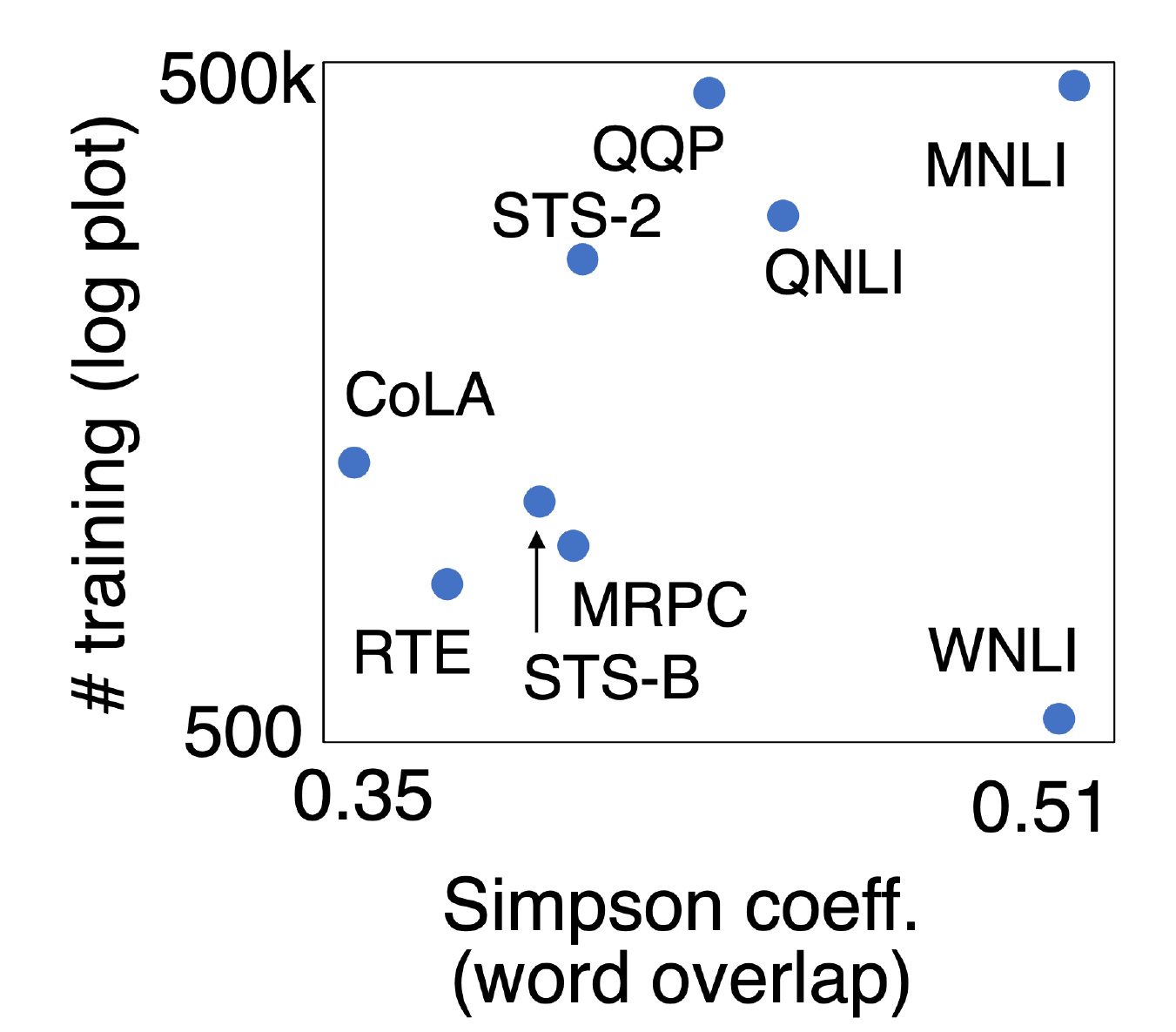}
\hspace{-3mm} \includegraphics[width=0.5\linewidth]{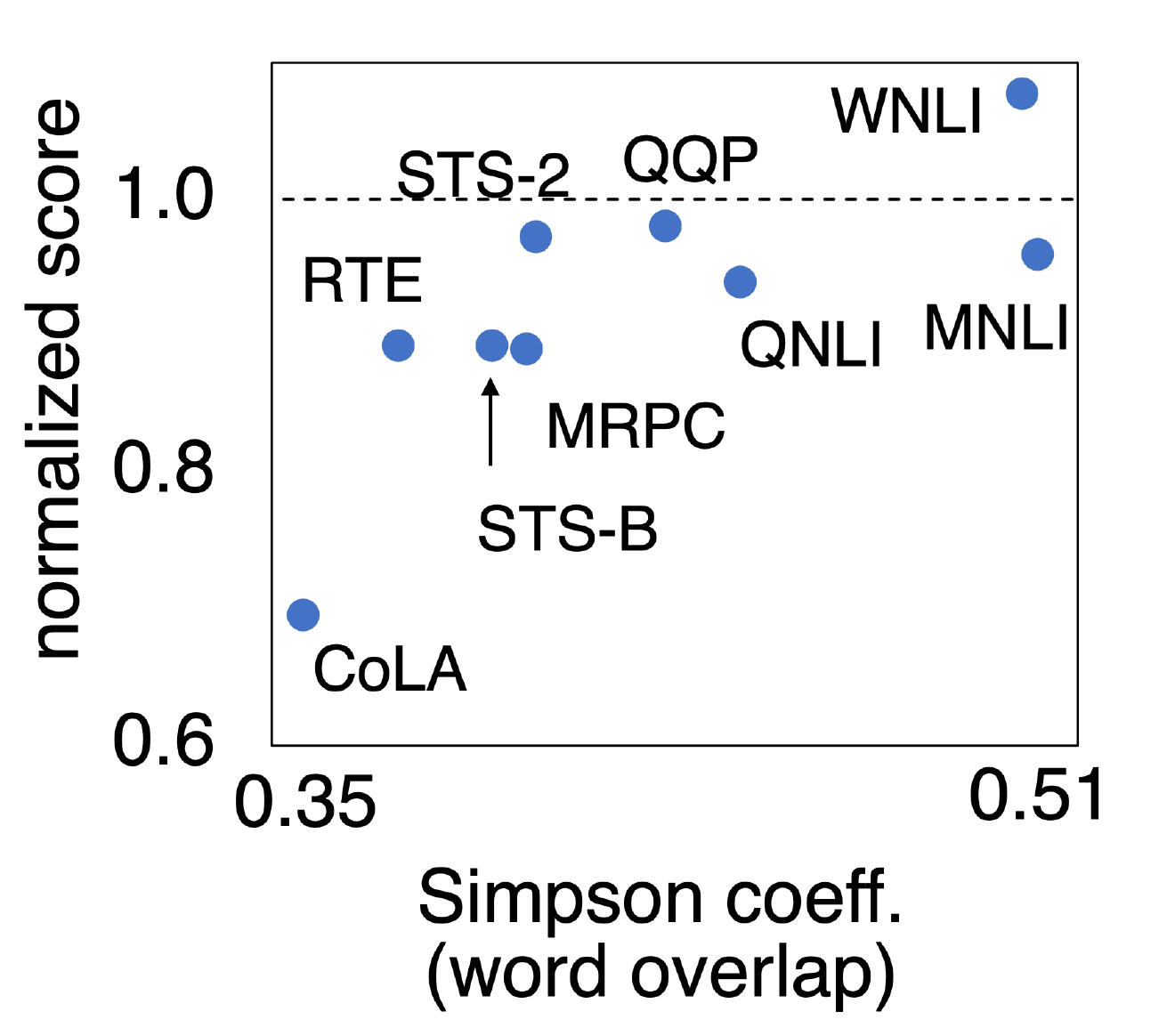}
\vspace{-2mm}
\caption{
Left: The number of training sentences vs. the Simpson coefficient between the GLUE and CC~(training) corpora. 
Right: The correlation between the Simpson coefficient and the model score.
The model scores were averaged over the five V\&L models and normalized with BERT$_{\rm BASE}$'s score.
\label{figure:task_scores}
}
\vspace{-4mm}
\end{figure}

\paragraph{Performance of each task.}
V\&L models perform lower than the BiLSTM baseline for some tasks, including MRPC, RTE, and WNLI.
Figure~\ref{figure:task_scores} (right) shows the correlation between the word overlap between the corpus for pre-training and the GLUE task corpora and the GLUE score.
We can see a positive correlation between those two variables.
Although we do not conclude clearly because word overlap and the number of training data also correlate, word overlap could have a large impact on task performance.

\section{Analysis}

\subsection{Amount of Change in Parameters}

\begin{table}[]
\small
\centering
\scalebox{0.94}{
\begin{tabular}{lcccc}
\hline
\textbf{} & \textbf{weight} & \textbf{weight (LN)} & \textbf{bias}& \textbf{bias (LN)} \\ \hline
\#layers         & 75 & 25 & 72 & 25 \\  \hline
VIS$_{\rm CTRL}$ & 0.9218 & 0.9999 & 0.9963 & 0.9973 \\
UNI$_{\rm CTRL}$ & 0.9197 & 0.9999 & 0.9966 & 0.9971 \\
VL$_{\rm CTRL}$  & 0.9193 & 0.9999 & 0.9964 & 0.9968 \\ \hdashline
VIL$_{\rm CTRL}$ & 0.9218 & 0.9999 & 0.9934 & 0.9895 \\
LX$_{\rm CTRL}$  & 0.9208 & 0.9998 & 0.9926 & 0.9935 \\ \hline
\end{tabular}
}
\vspace{-2mm}
\caption{Averaged cosine similarity between the corresponding parameters in the BERT$_{\rm BASE}$ and V\&L models. 
\#layers represents the number of layers transferred from BERT$_{\rm BASE}$ to V\&L models.
We computed the averaged similarity of the weights and biases in the layer normalization~(LN) layers and the other layers.
\label{table:parameters}
}
\vspace{-3mm}
\end{table}

We expected the model inference to be closer to BERT's inference if a model has parameters closer to BERT. Therefore, we calculated the cosine similarity of the corresponding parameters between pre-trained models and BERT to indicate the degree to which the parameters had changed.
Table~\ref{table:parameters} shows the averaged cosine similarity.
We flattened parameters and calculated their similarity as vectors.
We can see that the parameters of the single-stream and dual-stream models changed by the same extent.
This suggests that separating streams alone may not be sufficient for knowledge maintenance.

\subsection{Breakdown of Classification Results}
\label{sec:breakdown}

\begin{table}[]
\small
\centering
\begin{tabular}{lcccc}
\hline
 & \multicolumn{4}{c}{Successful models} \\
 & \textbf{Both} & \textbf{BERT}$_{\rm BASE}$ & \textbf{V\&L} & \textbf{Neither} \\ \hline
VIS$_{\rm CTRL}$ & 0.722 & 0.080 & 0.049 & 0.150 \\
UNI$_{\rm CTRL}$ & 0.717 & 0.085 & 0.050 & 0.149 \\
VL$_{\rm CTRL}$ & 0.700 & 0.102 & 0.053 & 0.146 \\ \hdashline
VIL$_{\rm CTRL}$ & 0.710 & 0.091 & 0.049 & 0.150 \\ 
LX$_{\rm CTRL}$ & 0.691 & 0.111 & 0.065 & 0.134 \\
\hline
\end{tabular}
\vspace{-2mm}
\caption{
Analysis of which models were successful in answering the classification task.
STS-B was excluded because it is a regression task.
We defined success in a problem as answering correctly in at least two out of three runs.
\label{table:breakdown}
}
\vspace{-3mm}
\end{table}

Table~\ref{table:breakdown} shows the results of aggregating the GLUE classification task problems into four categories: solvable by both BERT$_{\rm BASE}$ and V\&L models, BERT$_{\rm BASE}$ only, V\&L model only, and neither model.
We defined success in a given problem as answering correctly in at least two out of three experimental runs.
To make Table~\ref{table:breakdown}, we first calculated the tables of successful models for each GLUE task and V\&L model and second averaged the tables for the tasks.
For all five models, there are approximately 5\% of problems that they only can solve.
This category shows the positive impact of V\&L pre-training on NLU.
Problems that both models can solve tended to be more common for the single-stream models.
This supports the finding that these models retain more language knowledge.

The difference of corpora for the last pre-training between BERT (mainly English Wikipedia) and the V\&L models (images' alt-texts) might affect the complexity of the sentences in the problem sets that can be solved only by BERT and only by the V\&L models.
Thus, we analyzed the distributions of some metrics (sentence length, readability).
However, we found no significant difference between the two sets in each model. 
We show the distributions in Appendix~\ref{sec:appendix_analysis}.

\subsection{Language and V\&L Tasks}
\label{sec:l_and_vl}
The last column of Table~\ref{table:overall} shows the V\&L scores for the V\&L models.
We calculated these scores by averaging the results on the five V\&L tasks reported in \citet{bugliarello2020multimodal}.
Their tasks cover four groups widely used to test V\&L models: VQA, image--text retrieval, referring expressions, and multi-modal verification.
Comparing the V\&L and GLUE scores, we cans see that no model is best in both respects at the same time.
There is room for improvement in the V\&L extension.

\subsection{Structural Modification or Pre-training: Which Has the Greater Impact?}

\begin{table}[]
\small
\centering
\scalebox{0.94}{
\begin{tabular}{lcc}
\hline
\textbf{} & \textbf{Mod. only} & \textbf{Mod.+V\&L PT} \\ \hline
VIS$_{\rm CTRL}$ & 77.4 (1.00) & \textbf{72.5 (0.94)} \\
UNI$_{\rm CTRL}$ & 77.9 (1.01) & 71.4 (0.92) \\
VL$_{\rm CTRL}$ & 39.5 (0.51) & \textbf{72.4 (0.94)} \\ \hdashline
VIL$_{\rm CTRL}$ & 75.6 (0.98) & 70.9 (0.92) \\
LX$_{\rm CTRL}$ & \textbf{78.4 (1.01)} & 70.5 (0.91) \\ \hline
\end{tabular}
}
\vspace{-2mm}
\caption{Effect of V\&L pre-training on the averaged GLUE score. 
Values in parentheses are scores that have been normalized by the BERT$_{\rm BASE}$ scores.
\label{table:structure_mod_only}
}
\vspace{-3mm}
\end{table}

To further analyze the impact of structural modification, we fine-tuned models with only structural modifications (Mod. only).
Table~\ref{table:structure_mod_only} shows a comparison between the GLUE scores of the Mod-only models and the full models (Mod+V\&L-PT).
Except for VL$_{\rm CTRL}$, the Mod-only models achieve a score comparable to BERT$_{\rm BASE}$, and the GLUE score decreases for the Mod+V\&L-PT models.
The fact that the structural modification preserves the score of the GLUE tasks in most cases suggests that the main factor for the drop in the GLUE tasks is V\&L pre-training.
This observation emphasizes the impact of pre-training on maintaining the language knowledge.
Note that a possible reason for the exception of VL$_{\rm CTRL}$ is that the global visual feature added to the language embeddings may break the language knowledge.

\section{Discussion and Conclusion}

The number of V\&L model works that focus on both V\&L tasks and
language-only tasks has increased~\cite{ororbia2019like, lin2021m6v0,li2020unimo, hu2021transformer}.
\citet{ororbia2019like} proposed a V\&L neural architecture and trained it on a language model in a visual context. 
They demonstrated that their architecture outperforms its equivalent trained on language alone in perplexity and stated that language is inseparable from its physical context.
Although it is not clear whether methods that improve the perplexity of language modeling can also apply to maintain the performance of downstream tasks, the strategy of improving models with reference to human cognition would be an important direction.
More recently, \citet{li2020unimo} achieved better performance on language-only tasks than their base model with pre-training on three types of corpora (text, image, and image--text pairs) at the same time. 
\citet{lin2021m6v0} reported that adding separated extractors for vision and language on top of a single-stream encoder can help maintain language knowledge.

In this paper, we fine-tuned V\&L models extended from a language model (LM) to an NLU benchmark to compare their NLU performance.
We used five V\&L models, including single-stream and dual-stream models, pre-trained in the same setup.
The benchmark scores of those models decreased compared with their source LM. 
We also found that the single-stream models tended to retain (slightly) more language knowledge than the dual-stream models, and that the main cause of the drop in the NLU tasks can be pre-training.
Our observations suggest that adopting a single stream and devising pre-training strategies could be effective, at least for preserving the language knowledge.

\section*{Acknowledgements}
We would like to thank Takuma Udagawa, Taku Sakamoto, and the anonymous reviewers for their insightful comments.
This work was supported by JSPS KAKENHI Grant Number 21H03502.

\bibliography{anthology,custom}

\begin{thebibliography}{37}
\expandafter\ifx\csname natexlab\endcsname\relax\def\natexlab#1{#1}\fi

\bibitem[{Anderson et~al.(2018)Anderson, He, Buehler, Teney, Johnson, Gould,
  and Zhang}]{anderson2018bottom}
Peter Anderson, Xiaodong He, Chris Buehler, Damien Teney, Mark Johnson, Stephen
  Gould, and Lei Zhang. 2018.
\newblock \href
  {https://openaccess.thecvf.com/content_cvpr_2018/html/Anderson_Bottom-Up_and_Top-Down_CVPR_2018_paper.html}
  {{Bottom-up and top-down attention for image captioning and visual question
  answering}}.
\newblock In \emph{Proceedings of the IEEE Conference on Computer Vision and
  Pattern Recognition}, pages 6077--6086.

\bibitem[{Antol et~al.(2015)Antol, Agrawal, Lu, Mitchell, Batra, Zitnick, and
  Parikh}]{antol2015vqa}
Stanislaw Antol, Aishwarya Agrawal, Jiasen Lu, Margaret Mitchell, Dhruv Batra,
  C.~Lawrence Zitnick, and Devi Parikh. 2015.
\newblock \href
  {https://openaccess.thecvf.com/content_iccv_2015/html/Antol_VQA_Visual_Question_ICCV_2015_paper.html}
  {{VQA}: {V}isual {Q}uestion {A}nswering}.
\newblock In \emph{Proceedings of the IEEE International Conference on Computer
  Vision}, pages 2425--2433.

\bibitem[{Bar~Haim et~al.(2006)Bar~Haim, Dagan, Dolan, Ferro, Giampiccolo,
  Magnini, and Szpektor}]{bar2006second}
Roy Bar~Haim, Ido Dagan, Bill Dolan, Lisa Ferro, Danilo Giampiccolo, Bernardo
  Magnini, and Idan Szpektor. 2006.
\newblock \href {https://u.cs.biu.ac.il/~nlp/RTE2/Proceedings/01.pdf} {{The
  second {PASCAL} recognising textual entailment challenge}}.

\bibitem[{Bentivogli et~al.(2009)Bentivogli, Dagan, Dang, Giampiccolo, and
  Magnini}]{bentivogli2009fifth}
Luisa Bentivogli, Ido Dagan, Hoa~Trang Dang, Danilo Giampiccolo, and Bernardo
  Magnini. 2009.
\newblock \href
  {http://citeseerx.ist.psu.edu/viewdoc/summary?doi=10.1.1.232.1231} {{The
  Fifth {PASCAL} Recognizing Textual Entailment Challenge}}.
\newblock In \emph{Proceedings of the Text Analysis Conference (TAC’09)}.

\bibitem[{Bugliarello et~al.(2020)Bugliarello, Cotterell, Okazaki, and
  Elliott}]{bugliarello2020multimodal}
Emanuele Bugliarello, Ryan Cotterell, Naoaki Okazaki, and Desmond Elliott.
  2020.
\newblock \href {http://arxiv.org/abs/2011.15124} {{Multimodal Pretraining
  Unmasked: Unifying the Vision and Language BERTs}}.
\newblock \emph{arXiv preprint:2011.15124}.

\bibitem[{Cao et~al.(2020)Cao, Gan, Cheng, Yu, Chen, and Liu}]{cao2020behind}
Jize Cao, Zhe Gan, Yu~Cheng, Licheng Yu, Yen-Chun Chen, and Jingjing Liu. 2020.
\newblock \href
  {https://link.springer.com/chapter/10.1007/978-3-030-58539-6_34} {{Behind the
  scene: Revealing the secrets of pre-trained vision-and-language models}}.
\newblock In \emph{The 2020 European Conference on Computer Vision}, pages
  565--580.

\bibitem[{Cer et~al.(2017)Cer, Diab, Agirre, Lopez-Gazpio, and
  Specia}]{cer2017semeval}
Daniel Cer, Mona Diab, Eneko Agirre, I{\~n}igo Lopez-Gazpio, and Lucia Specia.
  2017.
\newblock \href {https://doi.org/10.18653/v1/S17-2001} {{{S}em{E}val-2017 Task
  1: Semantic Textual Similarity Multilingual and Crosslingual Focused
  Evaluation}}.
\newblock In \emph{Proceedings of the 11th International Workshop on Semantic
  Evaluation}, pages 1--14.

\bibitem[{Chen et~al.(2020)Chen, Li, Yu, Kholy, Ahmed, Gan, Cheng, and
  Liu}]{chen2020uniter}
Yen-Chun Chen, Linjie Li, Licheng Yu, Ahmed~El Kholy, Faisal Ahmed, Zhe Gan,
  Yu~Cheng, and Jingjing Liu. 2020.
\newblock \href {https://link.springer.com/chapter/10.1007/978-3-030-58577-8_7}
  {{UNITER}: Learning {UN}iversal image-{TE}xt representations}.
\newblock In \emph{The 2020 European Conference on Computer Vision}, pages
  104--120.

\bibitem[{Dagan et~al.(2005)Dagan, Glickman, and Magnini}]{dagan2006pascal}
Ido Dagan, Oren Glickman, and Bernardo Magnini. 2005.
\newblock \href {https://doi.org/10.1007/11736790_9} {{The {PASCAL} Recognising
  Textual Entailment Challenge}}.
\newblock In \emph{Proceedings of the First International Conference on Machine
  Learning Challenges: Evaluating Predictive Uncertainty Visual Object
  Classification, and Recognizing Textual Entailment}, page 177–190.

\bibitem[{Dolan and Brockett(2005)}]{dolan2005automatically}
William~B. Dolan and Chris Brockett. 2005.
\newblock \href {https://www.aclweb.org/anthology/I05-5002} {{Automatically
  Constructing a Corpus of Sentential Paraphrases}}.
\newblock In \emph{Proceedings of the Third International Workshop on
  Paraphrasing}.

\bibitem[{Giampiccolo et~al.(2007)Giampiccolo, Magnini, Dagan, and
  Dolan}]{giampiccolo2007third}
Danilo Giampiccolo, Bernardo Magnini, Ido Dagan, and Bill Dolan. 2007.
\newblock \href {https://www.aclweb.org/anthology/W07-1401} {{The Third
  {PASCAL} Recognizing Textual Entailment Challenge}}.
\newblock In \emph{Proceedings of the {ACL}-{PASCAL} Workshop on Textual
  Entailment and Paraphrasing}, pages 1--9.

\bibitem[{Haber et~al.(2019)Haber, Baumg{\"a}rtner, Takmaz, Gelderloos, Bruni,
  and Fern{\'a}ndez}]{haber2019photobook}
Janosch Haber, Tim Baumg{\"a}rtner, Ece Takmaz, Lieke Gelderloos, Elia Bruni,
  and Raquel Fern{\'a}ndez. 2019.
\newblock \href {https://doi.org/10.18653/v1/P19-1184} {{The {P}hoto{B}ook
  Dataset: Building Common Ground through Visually-Grounded Dialogue}}.
\newblock In \emph{Proceedings of the 57th Annual Meeting of the Association
  for Computational Linguistics}, pages 1895--1910.

\bibitem[{Hannan et~al.(2020)Hannan, Jain, and Bansal}]{hannan2020manymodalqa}
Darryl Hannan, Akshay Jain, and Mohit Bansal. 2020.
\newblock \href {https://ojs.aaai.org//index.php/AAAI/article/view/6294}
  {{ManyModalQA: Modality Disambiguation and QA over Diverse Inputs}}.
\newblock In \emph{Proceedings of the AAAI Conference on Artificial
  Intelligence}, volume~34, pages 7879--7886.

\bibitem[{Hu and Singh(2021)}]{hu2021transformer}
Ronghang Hu and Amanpreet Singh. 2021.
\newblock \href {http://arxiv.org/abs/2102.10772} {{Transformer is All You
  Need: Multimodal Multitask Learning with a Unified Transformer}}.
\newblock \emph{arXiv preprint arXiv:2003.13198}.

\bibitem[{Ilinykh et~al.(2019)Ilinykh, Zarrie{\ss}, and
  Schlangen}]{ilinykh2019meetup}
Nikolai Ilinykh, Sina Zarrie{\ss}, and David Schlangen. 2019.
\newblock \href {https://arxiv.org/abs/1907.05084} {{Meetup! a corpus of joint
  activity dialogues in a visual environment}}.
\newblock \emph{arXiv preprint arXiv:1907.05084}.

\bibitem[{Kazemzadeh et~al.(2014)Kazemzadeh, Ordonez, Matten, and
  Berg}]{KazemzadehOrdonezMattenBergEMNLP14}
Sahar Kazemzadeh, Vicente Ordonez, Mark Matten, and Tamara Berg. 2014.
\newblock \href {https://doi.org/10.3115/v1/D14-1086} {{{R}efer{I}t{G}ame:
  Referring to Objects in Photographs of Natural Scenes}}.
\newblock In \emph{Proceedings of the 2014 Conference on Empirical Methods in
  Natural Language Processing}, pages 787--798.

\bibitem[{Kembhavi et~al.(2017)Kembhavi, Seo, Schwenk, Choi, Farhadi, and
  Hajishirzi}]{kembhavi2017you}
Aniruddha Kembhavi, Minjoon Seo, Dustin Schwenk, Jonghyun Choi, Ali Farhadi,
  and Hannaneh Hajishirzi. 2017.
\newblock \href {https://ieeexplore.ieee.org/document/8100054} {{Are you
  smarter than a sixth grader? textbook question answering for multimodal
  machine comprehension}}.
\newblock In \emph{Proceedings of the IEEE Conference on Computer Vision and
  Pattern Recognition}, pages 4999--5007.

\bibitem[{Levesque et~al.(2012)Levesque, Davis, and
  Morgenstern}]{levesque2011winograd}
Hector~J. Levesque, Ernest Davis, and Leora Morgenstern. 2012.
\newblock \href {https://dl.acm.org/doi/10.5555/3031843.3031909} {{The Winograd
  Schema Challenge}}.
\newblock In \emph{Proceedings of the Thirteenth International Conference on
  Principles of Knowledge Representation and Reasoning}, page 552–561.

\bibitem[{Li et~al.(2019)Li, Yatskar, Yin, Hsieh, and Chang}]{li2019visualbert}
Liunian~Harold Li, Mark Yatskar, Da~Yin, Cho-Jui Hsieh, and Kai-Wei Chang.
  2019.
\newblock \href {https://arxiv.org/abs/1908.03557} {{VisualBERT: A simple and
  performant baseline for vision and language}}.
\newblock \emph{arXiv preprint arXiv:1908.03557}.

\bibitem[{Li et~al.(2020)Li, Gao, Niu, Xiao, Liu, Liu, Wu, and
  Wang}]{li2020unimo}
Wei Li, Can Gao, Guocheng Niu, Xinyan Xiao, Hao Liu, Jiachen Liu, Hua Wu, and
  Haifeng Wang. 2020.
\newblock \href {http://arxiv.org/abs/2012.15409} {{UNIMO: Towards
  Unified-Modal Understanding and Generation via Cross-Modal Contrastive
  Learning}}.
\newblock \emph{arXiv preprint arXiv:2012.15409}.

\bibitem[{Lin et~al.(2021)Lin, Yang, Zhang, Liu, Zhou, and Yang}]{lin2021m6v0}
Junyang Lin, An~Yang, Yichang Zhang, Jie Liu, Jingren Zhou, and Hongxia Yang.
  2021.
\newblock \href {http://arxiv.org/abs/2003.13198} {{M6-v0: Vision-and-Language
  Interaction for Multi-modal Pretraining}}.
\newblock \emph{arXiv preprint arXiv:2003.13198}.

\bibitem[{Lu et~al.(2019)Lu, Batra, Parikh, and Lee}]{NEURIPS2019_c74d97b0}
Jiasen Lu, Dhruv Batra, Devi Parikh, and Stefan Lee. 2019.
\newblock \href
  {https://papers.nips.cc/paper/2019/hash/c74d97b01eae257e44aa9d5bade97baf-Abstract.html}
  {{{ViLBERT}: Pretraining Task-Agnostic Visiolinguistic Representations for
  Vision-and-Language Tasks}}.
\newblock In \emph{Advances in Neural Information Processing Systems},
  volume~32.

\bibitem[{Nangia et~al.(2017)Nangia, Williams, Lazaridou, and
  Bowman}]{nangia2017repeval}
Nikita Nangia, Adina Williams, Angeliki Lazaridou, and Samuel Bowman. 2017.
\newblock \href {https://doi.org/10.18653/v1/W17-5301} {{The {R}ep{E}val 2017
  Shared Task: Multi-Genre Natural Language Inference with Sentence
  Representations}}.
\newblock In \emph{Proceedings of the 2nd Workshop on Evaluating Vector Space
  Representations for {NLP}}, pages 1--10.

\bibitem[{Ororbia et~al.(2019)Ororbia, Mali, Kelly, and
  Reitter}]{ororbia2019like}
Alexander Ororbia, Ankur Mali, Matthew Kelly, and David Reitter. 2019.
\newblock \href {https://aclanthology.org/P19-1506/} {Like a baby: Visually
  situated neural language acquisition}.
\newblock In \emph{Proceedings of the 57th Annual Meeting of the Association
  for Computational Linguistics}, pages 5127--5136.

\bibitem[{Rajpurkar et~al.(2016)Rajpurkar, Zhang, Lopyrev, and
  Liang}]{rajpurkar2016squad}
Pranav Rajpurkar, Jian Zhang, Konstantin Lopyrev, and Percy Liang. 2016.
\newblock \href {https://doi.org/10.18653/v1/D16-1264} {{{SQ}u{AD}: 100,000+
  Questions for Machine Comprehension of Text}}.
\newblock In \emph{Proceedings of the 2016 Conference on Empirical Methods in
  Natural Language Processing}, pages 2383--2392.

\bibitem[{Ren et~al.(2015)Ren, He, Girshick, and Sun}]{ren2016faster}
Shaoqing Ren, Kaiming He, Ross Girshick, and Jian Sun. 2015.
\newblock \href
  {https://proceedings.neurips.cc/paper/2015/file/14bfa6bb14875e45bba028a21ed38046-Paper.pdf}
  {{Faster R-CNN: Towards Real-Time Object Detection with Region Proposal
  Networks}}.
\newblock In \emph{Advances in Neural Information Processing Systems},
  volume~28.

\bibitem[{Sharma et~al.(2018)Sharma, Ding, Goodman, and
  Soricut}]{sharma-etal-2018-conceptual}
Piyush Sharma, Nan Ding, Sebastian Goodman, and Radu Soricut. 2018.
\newblock \href {https://doi.org/10.18653/v1/P18-1238} {{Conceptual Captions: A
  Cleaned, Hypernymed, Image Alt-text Dataset For Automatic Image Captioning}}.
\newblock In \emph{Proceedings of the 56th Annual Meeting of the Association
  for Computational Linguistics (Volume 1: Long Papers)}, pages 2556--2565.

\bibitem[{Socher et~al.(2013)Socher, Perelygin, Wu, Chuang, Manning, Ng, and
  Potts}]{socher2013recursive}
Richard Socher, Alex Perelygin, Jean Wu, Jason Chuang, Christopher~D. Manning,
  Andrew Ng, and Christopher Potts. 2013.
\newblock \href {https://www.aclweb.org/anthology/D13-1170} {{Recursive Deep
  Models for Semantic Compositionality Over a Sentiment Treebank}}.
\newblock In \emph{Proceedings of the 2013 Conference on Empirical Methods in
  Natural Language Processing}, pages 1631--1642.

\bibitem[{Su et~al.(2020)Su, Zhu, Cao, Li, Lu, Wei, and Dai}]{Su2020VL-BERT:}
Weijie Su, Xizhou Zhu, Yue Cao, Bin Li, Lewei Lu, Furu Wei, and Jifeng Dai.
  2020.
\newblock \href {https://openreview.net/forum?id=SygXPaEYvH} {{VL-BERT:
  Pre-training of Generic Visual-Linguistic Representations}}.
\newblock In \emph{International Conference on Learning Representations}.

\bibitem[{Suhr et~al.(2019)Suhr, Zhou, Zhang, Zhang, Bai, and
  Artzi}]{suhr-etal-2019-corpus}
Alane Suhr, Stephanie Zhou, Ally Zhang, Iris Zhang, Huajun Bai, and Yoav Artzi.
  2019.
\newblock \href {https://doi.org/10.18653/v1/P19-1644} {{A Corpus for Reasoning
  about Natural Language Grounded in Photographs}}.
\newblock In \emph{Proceedings of the 57th Annual Meeting of the Association
  for Computational Linguistics}, pages 6418--6428.

\bibitem[{Tan and Bansal(2019)}]{tan-bansal-2019-lxmert}
Hao Tan and Mohit Bansal. 2019.
\newblock \href {https://doi.org/10.18653/v1/D19-1514} {{{LXMERT}: Learning
  Cross-Modality Encoder Representations from Transformers}}.
\newblock In \emph{Proceedings of the 2019 Conference on Empirical Methods in
  Natural Language Processing and the 9th International Joint Conference on
  Natural Language Processing}, pages 5100--5111.

\bibitem[{Tanaka et~al.(2021)Tanaka, Nishida, and Yoshida}]{VisualMRC2021}
Ryota Tanaka, Kyosuke Nishida, and Sen Yoshida. 2021.
\newblock \href {https://arxiv.org/abs/2101.11272} {{VisualMRC: Machine Reading
  Comprehension on Document Images}}.
\newblock In \emph{the AAAI Conference on Artificial Intelligence}.

\bibitem[{Udagawa and Aizawa(2019)}]{udagawa2019natural}
Takuma Udagawa and Akiko Aizawa. 2019.
\newblock \href {https://ojs.aaai.org//index.php/AAAI/article/view/4694} {{A
  natural language corpus of common grounding under continuous and
  partially-observable context}}.
\newblock In \emph{Proceedings of the AAAI Conference on Artificial
  Intelligence}, volume~33, pages 7120--7127.

\bibitem[{Wang et~al.(2019)Wang, Singh, Michael, Hill, Levy, and
  Bowman}]{wang2018glue}
Alex Wang, Amanpreet Singh, Julian Michael, Felix Hill, Omer Levy, and
  Samuel~R. Bowman. 2019.
\newblock \href {https://openreview.net/forum?id=rJ4km2R5t7} {{GLUE: A
  Multi-Task Benchmark and Analysis Platform for Natural Language
  Understanding}}.
\newblock In \emph{International Conference on Learning Representations}.

\bibitem[{Warstadt et~al.(2019)Warstadt, Singh, and
  Bowman}]{warstadt-etal-2019-neural}
Alex Warstadt, Amanpreet Singh, and Samuel~R. Bowman. 2019.
\newblock \href {https://doi.org/10.1162/tacl_a_00290} {{Neural Network
  Acceptability Judgments}}.
\newblock \emph{Transactions of the Association for Computational Linguistics},
  7:625--641.

\bibitem[{Wolf et~al.(2020)Wolf, Debut, Sanh, Chaumond, Delangue, Moi, Cistac,
  Rault, Louf, Funtowicz, Davison, Shleifer, von Platen, Ma, Jernite, Plu, Xu,
  Le~Scao, Gugger, Drame, Lhoest, and Rush}]{wolf-etal-2020-transformers}
Thomas Wolf, Lysandre Debut, Victor Sanh, Julien Chaumond, Clement Delangue,
  Anthony Moi, Pierric Cistac, Tim Rault, Remi Louf, Morgan Funtowicz, Joe
  Davison, Sam Shleifer, Patrick von Platen, Clara Ma, Yacine Jernite, Julien
  Plu, Canwen Xu, Teven Le~Scao, Sylvain Gugger, Mariama Drame, Quentin Lhoest,
  and Alexander Rush. 2020.
\newblock \href {https://doi.org/10.18653/v1/2020.emnlp-demos.6}
  {{Transformers: State-of-the-Art Natural Language Processing}}.
\newblock In \emph{Proceedings of the 2020 Conference on Empirical Methods in
  Natural Language Processing: System Demonstrations}, pages 38--45.

\bibitem[{Yagcioglu et~al.(2018)Yagcioglu, Erdem, Erdem, and
  Ikizler-Cinbis}]{yagcioglu-etal-2018-recipeqa}
Semih Yagcioglu, Aykut Erdem, Erkut Erdem, and Nazli Ikizler-Cinbis. 2018.
\newblock \href {https://doi.org/10.18653/v1/D18-1166} {{{R}ecipe{QA}: A
  Challenge Dataset for Multimodal Comprehension of Cooking Recipes}}.
\newblock In \emph{Proceedings of the 2018 Conference on Empirical Methods in
  Natural Language Processing}, pages 1358--1368.

\end{thebibliography}
\bibliographystyle{acl_natbib}

\appendix
\section{Dataset Statistics}
\label{sec:appendix}

\begin{table}[h!]
\small
\centering
\scalebox{0.94}{
\begin{tabular}{lccccc}
\hline
\textbf{Dataset} & \textbf{Task} & \textbf{Size} & \textbf{\#vocab.} & \begin{tabular}[c]{@{}c@{}} \textbf{Word ov} \\ \textbf{btw CC} \end{tabular} \\ \hline
\multicolumn{5}{c}{V\&L pre-training} \\
CC & CAP & 2.8M & 48,360 & 1 \\
CC (val) & CAP & 14K & 10,442 & 0.63 \\ \hline
\multicolumn{5}{c}{GLUE benchmark} \\
WNLI & NLI & 635 & 1.622 & 0.08 \\
RTE & NLI & 2.5K & 23,341 & 0.24 \\
MRPC & P/S & 3.7K & 13,926 & 0.26 \\
STS-B & P/S & 5.7K & 16,436 & 0.25 \\
CoLA & SS & 8.6K & 7,845 & 0.19 \\
SST-2 & SS & 67K & 14,816 & 0.26 \\
QNLI & NLI & 104K & 148,413 & 0.29 \\
QQP & P/S & 364K & 193,041 & 0.28 \\
MNLI & NLI & 393K & 167,790 & 0.34 \\ \hline
\end{tabular}
}
\vspace{-2mm}
\caption{Training dataset statistics. 
CC: The Conceptual Captions dataset \cite{sharma-etal-2018-conceptual}.
CAP: image captioning, P/S: paraphrase/similarity task, SS: single-sentence task.
\label{table:glue}
}
\vspace{-3mm}
\end{table}

\section{Additional Data for Overall Results}
\label{sec:appendix_overall}

Table~\ref{table:overall.sd} shows the SDs to the averaged scores of V\&L models on the GLUE tasks' development sets.

\begin{table}[h!]
\small
\centering
\scalebox{0.94}{
\begin{tabular}{l}
\begin{tabular}{lccc}
\hline
\textbf{} & \textbf{avg} & \textbf{CoLA} & \textbf{SST-2} \\ \hline

BERT$_{\rm BASE}$ & 
\begin{tabular}[c]{@{}c@{}} 77.3 \\ (0.8) \end{tabular} &
\begin{tabular}[c]{@{}c@{}} 54.6 \\ (1.1) \end{tabular} &
\begin{tabular}[c]{@{}c@{}} 92.5 \\ (0.1) \end{tabular} \\ \hline

VIS$_{\rm CTRL}$ & 
\begin{tabular}[c]{@{}c@{}} 72.5 \\ (1.2) \end{tabular} &
\begin{tabular}[c]{@{}c@{}} 38.6 \\ (7.3) \end{tabular} &
\begin{tabular}[c]{@{}c@{}} 89.4 \\ (0.4) \end{tabular} \\

UNI$_{\rm CTRL}$ & 
\begin{tabular}[c]{@{}c@{}} 71.4 \\ (0.3) \end{tabular} &
\begin{tabular}[c]{@{}c@{}} 37.4 \\ (6.5) \end{tabular} &
\begin{tabular}[c]{@{}c@{}} 89.7 \\ (0.5) \end{tabular} \\

VL$_{\rm CTRL}$& 
\begin{tabular}[c]{@{}c@{}} 72.4 \\ (0.8) \end{tabular} &
\begin{tabular}[c]{@{}c@{}} 38.7 \\ (1.5) \end{tabular} &
\begin{tabular}[c]{@{}c@{}} 89.8 \\ (0.9) \end{tabular} \\ \hdashline

VIL$_{\rm CTRL}$ & 
\begin{tabular}[c]{@{}c@{}} 70.9 \\ (0.8) \end{tabular} &
\begin{tabular}[c]{@{}c@{}} 36.1 \\ (6.0) \end{tabular} &
\begin{tabular}[c]{@{}c@{}} 90.4 \\ (0.5) \end{tabular} \\

LX$_{\rm CTRL}$ & 
\begin{tabular}[c]{@{}c@{}} 70.5 \\ (0.2) \end{tabular} &
\begin{tabular}[c]{@{}c@{}} 39.0 \\ (6.1) \end{tabular} &
\begin{tabular}[c]{@{}c@{}} 90.2 \\ (0.5) \end{tabular} \\ \hline

\end{tabular} \\
\\

\begin{tabular}{lccc}
\hline
\textbf{} & \textbf{MRPC} & \textbf{QQP} & \textbf{STS-B} \\ \hline

BERT$_{\rm BASE}$ & 
\begin{tabular}[c]{@{}c@{}} 81.9 / 87.6 \\ (0.6) / (0.5) \end{tabular} &
\begin{tabular}[c]{@{}c@{}} 90.6 / 87.4 \\ (0.0) / (0.1) \end{tabular} &
\begin{tabular}[c]{@{}c@{}} 88.2 / 87.9 \\ (0.3) / (0.3) \end{tabular} \\ \hline

VIS$_{\rm CTRL}$ & 
\begin{tabular}[c]{@{}c@{}} 71.9 / 82.1 \\ (1.4) / (0.8) \end{tabular} &
\begin{tabular}[c]{@{}c@{}} 89.4 / 86.0 \\ (0.1) / (0.1) \end{tabular} &
\begin{tabular}[c]{@{}c@{}} 81.8 / 81.7 \\ (4.0) / (3.6) \end{tabular} \\

UNI$_{\rm CTRL}$ & 
\begin{tabular}[c]{@{}c@{}} 74.9 / 75.6 \\ (2.0) / (2.2) \end{tabular} &
\begin{tabular}[c]{@{}c@{}} 69.3 / 80.3 \\ (0.8) / (0.7) \end{tabular} &
\begin{tabular}[c]{@{}c@{}} 89.2 / 85.7 \\ (0.1) / (0.1) \end{tabular} \\

VL$_{\rm CTRL}$ & 
\begin{tabular}[c]{@{}c@{}} 70.6 / 81.8 \\ (0.5) / (0.3) \end{tabular} &
\begin{tabular}[c]{@{}c@{}} 89.0 / 85.4 \\ (0.3) / (0.4) \end{tabular} &
\begin{tabular}[c]{@{}c@{}} 82.9 / 82.8 \\ (2.3) / (1.9) \end{tabular} \\ \hdashline

VIL$_{\rm CTRL}$ & 
\begin{tabular}[c]{@{}c@{}} 69.0 / 79.4 \\ (1.3) / (2.1) \end{tabular} &
\begin{tabular}[c]{@{}c@{}} 88.6 / 85.0 \\ (0.2) / (0.1) \end{tabular} &
\begin{tabular}[c]{@{}c@{}} 77.7 / 78.0 \\ (1.2) / (0.9) \end{tabular} \\

LX$_{\rm CTRL}$ & 
\begin{tabular}[c]{@{}c@{}} 69.8 / 80.4 \\ (1.3) / (1.1) \end{tabular} &
\begin{tabular}[c]{@{}c@{}} 89.0 / 85.4 \\ (0.1) / (0.2) \end{tabular} &
\begin{tabular}[c]{@{}c@{}} 75.3 / 75.3 \\ (0.8) / (0.7) \end{tabular} \\ \hline

\end{tabular} \\
\\

\begin{tabular}{lcccc}
\hline
\textbf{} & \textbf{MNLI} & \textbf{QNLI} & \textbf{RTE} & \textbf{WNLI} \\ \hline

BERT$_{\rm BASE}$ & 
\begin{tabular}[c]{@{}c@{}} 84.2 \\ (0.1) \end{tabular} &
\begin{tabular}[c]{@{}c@{}} 91.0 \\ (0.4) \end{tabular} &
\begin{tabular}[c]{@{}c@{}} 62.5 \\ (1.5) \end{tabular} &
\begin{tabular}[c]{@{}c@{}} 48.8 \\ (5.8) \end{tabular} \\ \hline

VIS$_{\rm CTRL}$ & 
\begin{tabular}[c]{@{}c@{}} 81.6 \\ (0.2) \end{tabular} &
\begin{tabular}[c]{@{}c@{}} 87.0 \\ (1.1) \end{tabular} &
\begin{tabular}[c]{@{}c@{}} 56.6 \\ (1.9) \end{tabular} &
\begin{tabular}[c]{@{}c@{}} 53.1 \\ (4.6) \end{tabular} \\

UNI$_{\rm CTRL}$ & 
\begin{tabular}[c]{@{}c@{}} 80.9 \\ (0.4) \end{tabular} &
\begin{tabular}[c]{@{}c@{}} 86.0 \\ (1.0) \end{tabular} &
\begin{tabular}[c]{@{}c@{}} 55.6 \\ (2.4) \end{tabular} &
\begin{tabular}[c]{@{}c@{}} 55.4 \\ (1.3) \end{tabular} \\

VL$_{\rm CTRL}$& 
\begin{tabular}[c]{@{}c@{}} 81.2 \\ (0.2) \end{tabular} &
\begin{tabular}[c]{@{}c@{}} 86.3 \\ (0.1) \end{tabular} &
\begin{tabular}[c]{@{}c@{}} 55.7 \\ (1.4) \end{tabular} &
\begin{tabular}[c]{@{}c@{}} 53.1 \\ (3.5) \end{tabular} \\ \hdashline

VIL$_{\rm CTRL}$ & 
\begin{tabular}[c]{@{}c@{}} 79.9 \\ (0.5) \end{tabular} &
\begin{tabular}[c]{@{}c@{}} 83.8 \\ (0.6) \end{tabular} &
\begin{tabular}[c]{@{}c@{}} 53.7 \\ (0.9) \end{tabular} &
\begin{tabular}[c]{@{}c@{}} 55.4 \\ (1.8) \end{tabular} \\

LX$_{\rm CTRL}$ & 
\begin{tabular}[c]{@{}c@{}} 80.4 \\ (0.2) \end{tabular} &
\begin{tabular}[c]{@{}c@{}} 84.2 \\ (0.2) \end{tabular} &
\begin{tabular}[c]{@{}c@{}} 57.2 \\ (3.4) \end{tabular} &
\begin{tabular}[c]{@{}c@{}} 46.0 \\ (9.2) \end{tabular} \\ \hline

\end{tabular}

\end{tabular}
}
\caption{
Standard deviations of our results in the performance on the GLUE tasks' development sets (Table~\ref{table:overall}).
SDs are shown in parentheses below each value.
We ran three experiments for each task. 
\label{table:overall.sd}
}
\end{table}

\section{Additional Data for Analysis}
\label{sec:appendix_analysis}

\begin{figure}[]
\centering
\includegraphics[width=\linewidth]{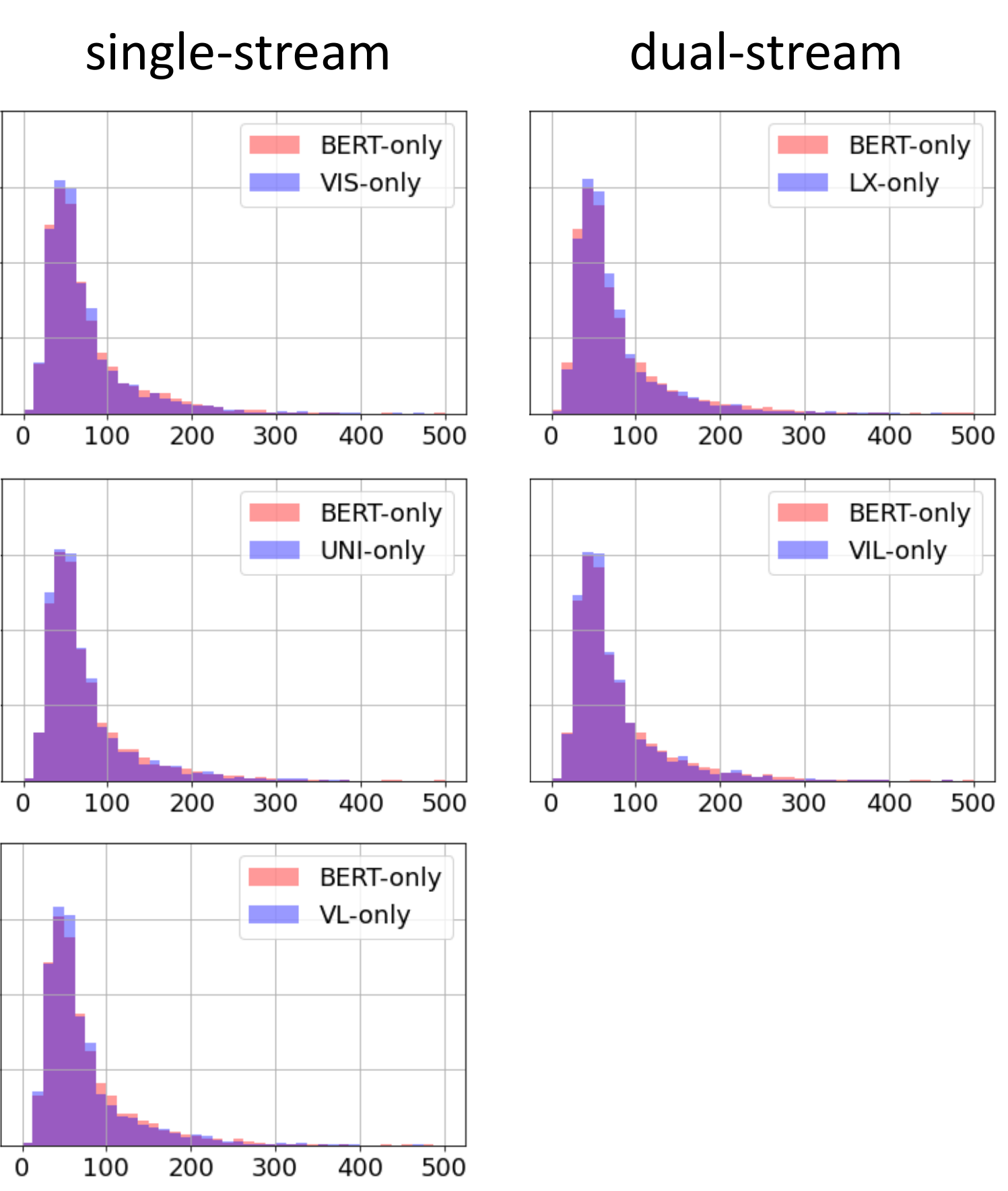}
\caption{
The sentence length distributions in the problem sets solved only by the V\&L model and only by BERT.
In each plot, the area of the distribution is normalized to 1.
The range of the vertical axis is [0, 0.020].
\label{figure:length}
}
\vspace{-3mm}
\end{figure}

\begin{figure}[]
\centering
\includegraphics[width=\linewidth]{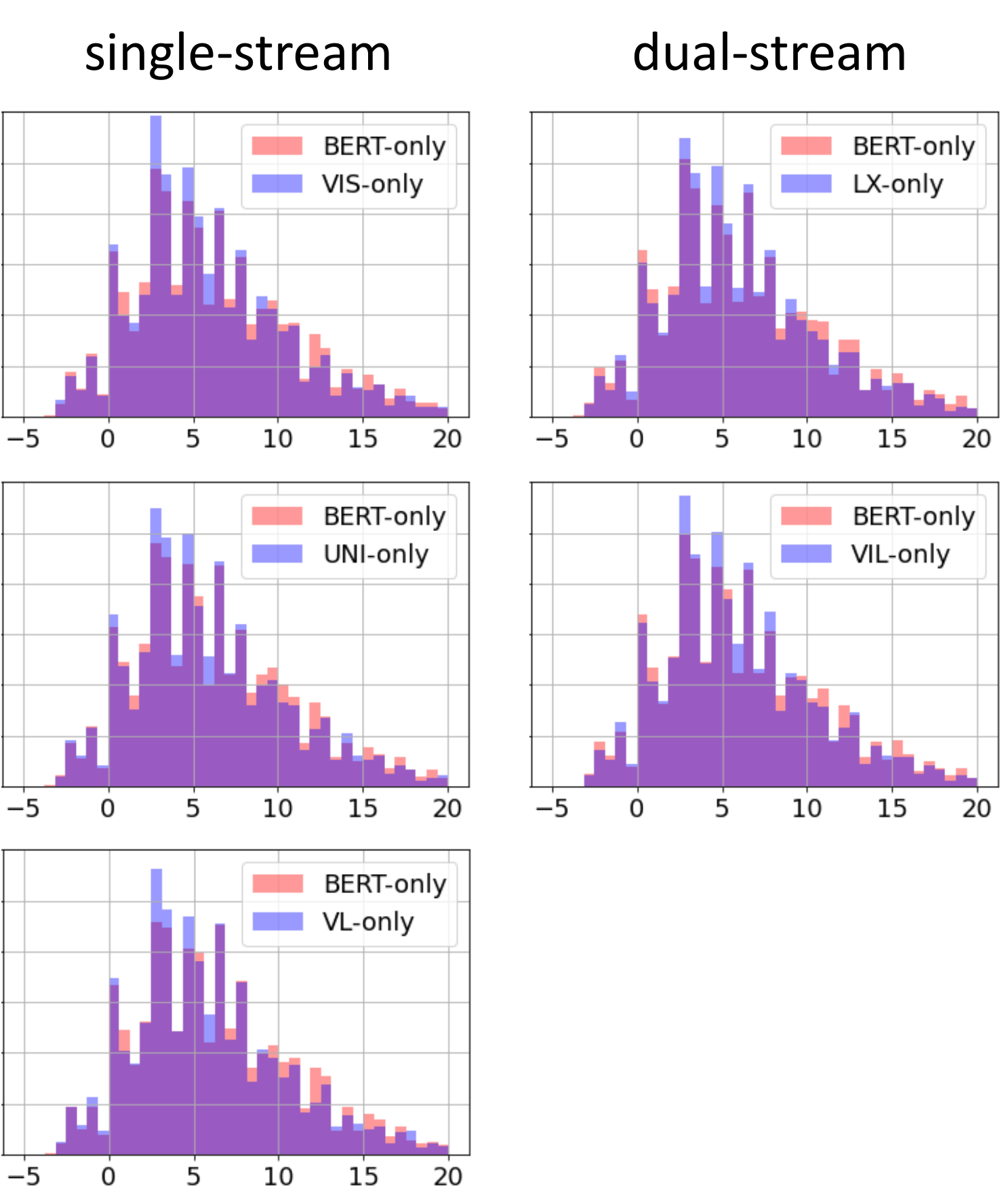}
\caption{
The Flesch--Kincaid Grade Level distributions of sentences in the problem sets solved only by the V\&L model and only by BERT.
In each plot, the area of the distribution is normalized to 1.
The range of the vertical axis is [0, 0.15].
\label{figure:readablity}
}
\vspace{-3mm}
\end{figure}

We show the distributions of sentence length and readability mentioned in Section~\ref{sec:breakdown} in Figure~\ref{figure:length} and Figure~\ref{figure:readablity}, respectively.

\end{document}